\documentclass[10pt,twocolumn,letterpaper]{article}

\usepackage{cvpr}
\usepackage{times}
\usepackage{epsfig}
\usepackage{graphicx}
\usepackage{amsmath}
\usepackage{amssymb}
\usepackage{bm}
\usepackage{color}
\usepackage{booktabs}
\usepackage{algorithm}
\usepackage{algorithmic}
\usepackage{multirow}
\usepackage[numbers,sort&compress]{natbib}

\usepackage{xcolor}
\definecolor{green}{rgb}{0.0, 0.6, 0.0}
\hypersetup{pagebackref=false,breaklinks=true,letterpaper=true,colorlinks,bookmarks=false}

\cvprfinalcopy 

\def\cvprPaperID{2434} 

\ifcvprfinal\fi
\begin{document}

\title{Boosting Few-Shot Learning With Adaptive Margin Loss}

\author{Aoxue Li$^{1}$\thanks{This work was done when the first author was an intern at Huawei Noah's Ark Lab.} \quad Weiran Huang$^{2}$ \quad Xu Lan$^{3}$ \quad Jiashi Feng$^{4}$ \quad Zhenguo Li$^{2}$ \quad Liwei Wang$^{1}$\vspace{0.5em}\\
  $^1$School of EECS, Peking University, China\\
  $^2$Huawei Noah's Ark Lab, China\\
  $^3$Queen Mary University of London, UK\\
  $^4$National University of Singapore, Singapore\\
  {\tt\small lax@pku.edu.cn, weiran.huang@outlook.com, x.lan@qmul.ac.uk,}\\ {\tt\small elefjia@nus.edu.sg, li.zhenguo@huawei.com, wanglw@cis.pku.edu.cn}
}

\maketitle
\thispagestyle{empty}

\begin{abstract}
Few-shot learning (FSL) has attracted increasing attention in recent years but remains challenging, due to the intrinsic difficulty in learning to generalize from a few examples. This paper proposes an adaptive margin principle to improve the generalization ability of metric-based meta-learning approaches for few-shot learning problems. Specifically, we first develop a class-relevant additive margin loss, where semantic similarity between each pair of classes is considered to separate samples in the feature embedding space from similar classes. Further, we incorporate the semantic context among all classes in a sampled training task and develop a task-relevant additive margin loss to better distinguish samples from different classes. Our adaptive margin method can be easily extended to a more realistic generalized FSL setting. Extensive experiments demonstrate that the proposed method can boost the performance of current metric-based meta-learning approaches, under both the standard FSL and generalized FSL settings.
\end{abstract}

\section{Introduction}

Deep learning has achieved great success in various computer vision tasks~\cite{Russakovsky2015ImageNet,he2016cvpr}. However, with a large number of parameters, deep neural networks require large amounts of labeled data for model training. This severely limits their scalability -- for many rare classes, it is infeasible to collect a large number of labeled samples. In contrast, humans can recognize an object after seeing it once. Inspired by the few-shot learning ability of humans, there has been an increasing interest in the few-shot learning (FSL) problem~\cite{Ren2018arxiv,Snell2017nips,Douze2018cvpr,Li2019CVPR}. Given a set of base classes with sufficient labeled samples, and a set of novel classes with only a few labeled samples, FSL aims to learn a classifier for the novel classes by learning a generic knowledge from the base classes.

\begin{figure}[t]
\begin{center}
\includegraphics[width=0.9\columnwidth]{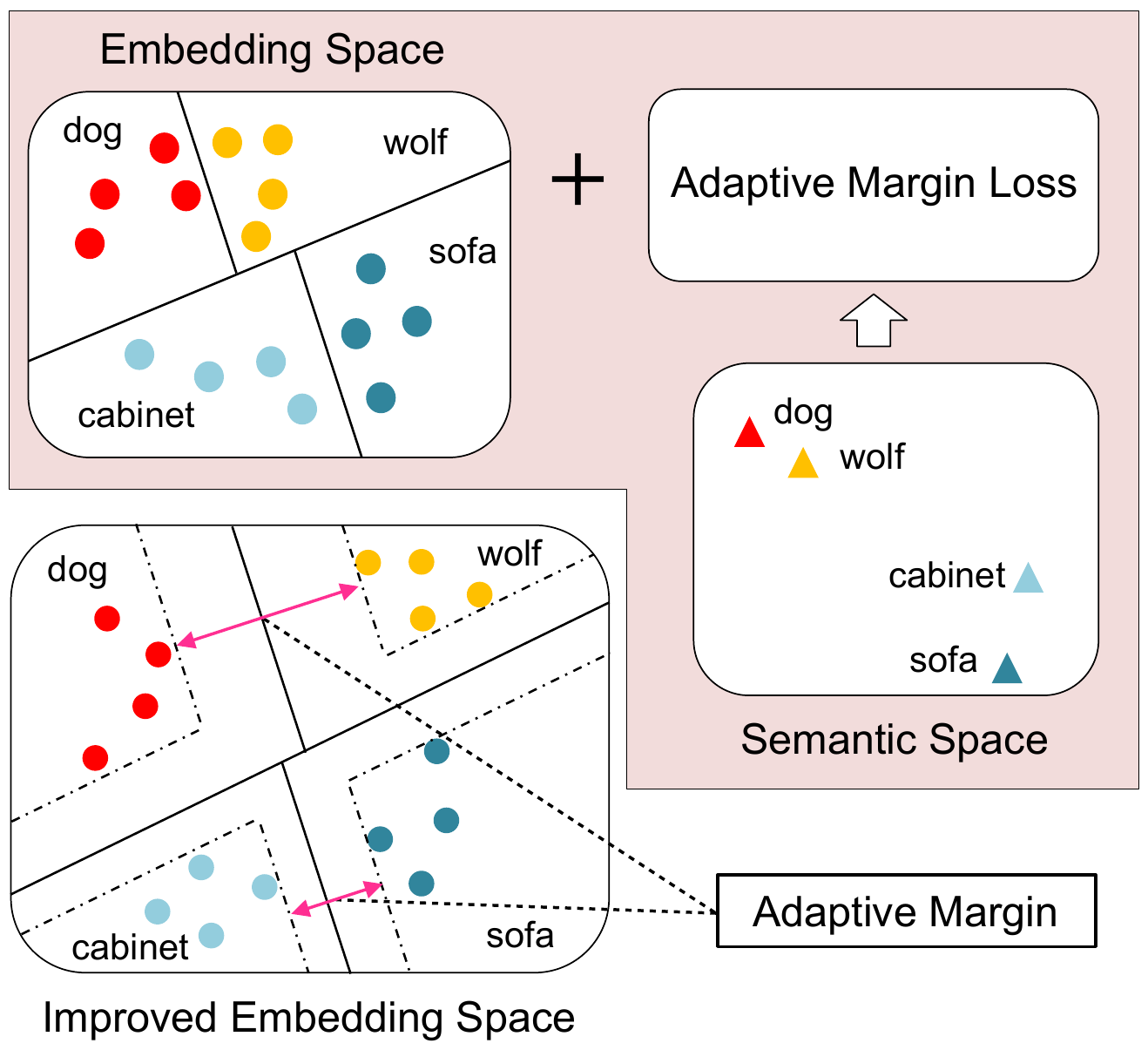}
\end{center}
\caption{The illustration of the key insight of our adaptive margin loss. In our approach, semantic similarities between different classes (measured in the semantic space of classes) are leveraged to generate adaptive margin between classes. Then, the margin is integrated into the classification loss to make similar classes more separable in the embedding space, which benefits FSL.}
\label{fig0}
\end{figure}

Recently, metric-based meta-learning approaches~\cite{Snell2017nips,Li2017arxiv,Gidaris2018cvpr,Sung2018cvpr,Li2019ICCV} have shown the superior performance in solving the FSL problem, with attractive simplicity. These approaches usually learn a good embedding space, where samples from the same class are clustered together while samples from different classes are far away from each other. 
In this way, a new sample from the novel class can be recognized directly through a simple distance metric within the learned embedding space. The success of these metric-based approaches relies on learning a discriminative embedding space. 

To further improve the performance, we introduce the adaptive margin in the embedding space, 
which helps to separate samples from different classes, especially for similar classes. 
The key insight of our approach is that the semantic similarity between different classes can be leveraged to generate adaptive margin between classes, \ie, the margin between similar classes should be larger than the one between dissimilar classes (as illustrated in Figure~\ref{fig0}). By integrating the adaptive margin into the classification loss, our method learns a more discriminative embedding space with better generalization ability.

Specifically, we first propose a class-relevant margin generator which produces an adaptive margin for each pair of classes based on their semantic similarity in the semantic space. By combining the margin generated by class-relevant margin generator and the classification loss of FSL approaches, our class-relevant additive margin loss can effectively pull each class away from other classes. Considering the semantic context among a sampled training task in the FSL, we further develop a task-relevant margin generator. By comparing each class with the rest classes among the task in the semantic space, our task-relevant margin generator produces more suitable margin for each pair of classes. By involving these margin penalty, our task-relevant margin loss learns more discriminative embedding space, thus leads to stronger generalization ability to recognize novel class samples.
Moreover, our approach can be easily extended to a more realistic yet more challenging FSL setting (\ie, the generalized FSL) where the label space of test data covers both base and novel classes. This is as opposed to the standard FSL setting where the test data contain novel class samples only. Experimental results on the two FSL benchmarks show that our approach significantly improves the performance of current metric-learning-based approaches on both of the two FSL settings.

In summary, our contributions are three folds: (1) To the best of our knowledge, this is the first work to propose an adaptive margin principle to improve the performance of current metric-based meta-learning approaches for FSL. (2) We propose a task-relevant adaptive margin loss to well distinguish samples from different classes in the embedding space according to their semantic similarity, and experimental results demonstrate that our method achieves the state-of-the-art results on the benchmark dataset. (3) Our approach can be easily extended to a more realistic yet more challenging generalized FSL setting, with superior performance obtained. This further validates the effectiveness of our method.

\section{Related Work}

\subsection{Few-shot Learning}

In recent years, few-shot object recognition has become topical. With the success of deep convolutional neural network (DCNN) based approaches in the data-rich setting~\cite{Russakovsky2015ImageNet,he2016cvpr,szegedy2015cvpr,donahue2014decaf}, there has been
a great of interest in generalizing such deep learning approaches to the few-shot setting. Most of the recent approaches use a meta-learning strategy. With the meta-learning, these models extract transferable knowledge from a set of auxiliary tasks via episodic training. The knowledge then helps  to learn the few-shot classifier trained for the novel classes. 

Existing meta-learning based FSL approaches usually learn a model that, given a task (a set of few-shot labeled data and some test query data), produces a classifier that generalizes across all tasks~\cite{Gidaris2018cvpr}. A main group of gradient-based meta-learning models attempt to modify the classical gradient-based optimization to adapt to a new episodic task by producing efficient parameter updates~\cite{Finn2017icml,Gidaris2018cvpr,Antoniou2018iclr,Ravi2017iclr}. Recently, many meta-learning approaches attempt to learn an effective metric on the feature space. The intuition is that if a model can determine the similarity of two images, it can classify an unseen test image with a few labeled examples~\cite{Snell2017nips,Sung2018cvpr}. To learn an effective metric, these methods make their prediction conditioned on distances to a few labeled examples during the training stage~\cite{Bertinetto2019iclr,vinyals2016bnips}. These examples are sampled from base classes designed to simulate the few-shot scenario. In this paper, we propose a novel generic adaptive margin strategy which can be integrated in existing metric-based meta-learning approaches. Our method can force different classes far from each other in the embedding space. This makes it much easier to recognize novel class samples. 

\subsection{Margin Loss in Visual Recognition}

Softmax loss has been widely used in training DCNNs for extracting discriminative visual features for object recognition tasks. By observing that the weights from the last fully connected layer of a classification DCNN trained on the softmax loss bear conceptual similarities with the centers of each class, the works in \cite{Deng2019cvpr,Wang2018cvprCosface,Liu2017cvprsphere} proposed several margin losses to improve the discriminative power of the trained model. Liu~\etal~\cite{Liu2017cvprsphere} introduced the important idea of angular margin. However, their loss function required a series of approximations in order to be computed, which resulted in an unstable training of the network. Wang~\etal~\cite{Wang2018iclrws} and Wang~\etal~\cite{Wang2018cvprCosface} directly add cosine margin to the target logits and achieve better results than~\cite{Liu2017cvprsphere}. Deng~\etal~\cite{Deng2019cvpr} proposed an additive angular margin loss to further improve the discriminative power of feature embedding space. Although the aforementioned margin losses have achieved promising results on visual recognition tasks, they are not designed for FSL, where limited samples are provided for novel classes. To learn more suitable margin for FSL, we thus propose an adaptive margin principle, where the semantic context among a sampled training task is considered. By training the FSL approach with our adaptive margin loss, the learned model generalizes better across all tasks and thus achieves better recognition results on novel classes.

\section{Methodology}

\subsection{Preliminary: Metric-Based Meta-Learning}

In the few-shot learning (FSL), we are given a base class set $C_{base}$ consisting of $n_{base}$ base classes, and for each base class, we have sufficient labeled samples.
Meanwhile, we also have a novel class set $C_{novel}$ with $n_{novel}$ novel classes, each of which has only a few labeled
samples (\eg, less than 5 samples).
The {\bf goal} of FSL is to obtain a good classifier for the novel classes. 

Meta-learning~\cite{Wang2018cvprlsl,Snell2017nips,vinyals2016bnips,Finn2017icml,Chen2019NIPS} is a common approach for the FSL.
A standard meta-learning procedure involves two stages: meta-training and meta-testing. 
In the meta-training stage, we train the model in an episodic manner. In each episode, a small classification task is constructed by sampling a small training set and a small test set from the whole base class dataset, and then it is used to update the model.
In the meta-testing stage, the learned model is used to recognize samples from novel classes.
Recently, metric-based meta-learning approaches become popular \cite{Chen2019NIPS,Gidaris2018cvpr}. 
Most metric-based meta-learning approaches generally assume that there exists an embedding space in which samples cluster around a single representation for each class, and
then these class representations are used as references to infer labels of test samples. In the following, we introduce the framework of metric-based meta-learning approaches.

\noindent\textbf{Meta-Training.} 
In each episode of meta-training, we sample a $n_t$-way $n_s$-shot classification task from the base class dataset. Specifically, we randomly choose $n_t$ classes from base class set $C_{base}$ for the episodic training, denoted as $C_{t}$. We randomly select $n_s$ samples from each episodic training class and combine them to form a small training set, which is called support set $S$. Moreover, we also randomly select some other samples from each episodic training class and combine them to form a small test set, which is called query set $Q$.

In the current episode, all samples from both query set and support set are embedded into the embedding space by using an embedding module $\mathcal{F}$. Then, the meta-learner generates class representations 
$r_1, r_2, \cdots, r_{n_t}$ by using the samples from support set $S$. 
For example, Prototypical Networks \cite{Snell2017nips} generates class representations by averaging the embeddings of support samples by class. After that, the meta-learner uses a metric module $\mathcal{D}$ (\eg, cosine similarity) to measure the similarity between every query point $(x,y) \in Q$ and the current class representations in the embedding space. Based on these similarities, the meta-learner incurs a classification loss for each point in the current query set. The meta-learner then back-propagates the gradient of the total loss of all query samples. The classification loss can be formulated as:
\begin{equation}
\mathcal{L}^{\rm cls}=- \frac{1}{|Q|}\sum_{(x,y)\in Q}\log\frac{e^{\mathcal{D}(\mathcal{F}(x),r_y)}}{\sum\limits_{k\in C_{t}} e^{\mathcal{D}(\mathcal{F}(x),r_k)}},
\label{softmax}
\end{equation}
where $\mathcal{D}(\mathcal{F}(x),r_k)$ denotes the similarity between sample $x$ and the $k$-th class representation $r_k$ predicted by the meta-learner.

\noindent\textbf{Meta-Testing.}  In an episode of meta-testing, a novel classification task is similar to a training base classification task. Specifically, the labeled few-shot sample set and unlabeled test examples are used to form the support set and query set, respectively. Then they are fed into the learned model with predicted classification results of query samples as outputs.

Different metric-based meta-learning  approaches differ in the form of the class representation generation module and metric module, our work introduces different margin loss to improve current metric-based meta-learning approaches.

\subsection{Naive Additive Margin Loss}
\label{sect:simple_additive}

An intuitive idea to learn a discriminative embedding space is to add a  margin between the predicted results of different classes. This helps to increase the inter-class distance in the embedding space and make it easier to recognize test novel samples.
To achieve this, we propose a naive additive margin loss (NAML), which can be formulated as:
\begin{equation}
\small
\mathcal{L}^{\rm na}\! =\!- \frac{1}{|Q|}\sum_{(x,y)\in Q}\log p^{\rm na}(y|x,S),
\label{naive_additive_margin_loss}
\end{equation}
where 
\begin{flalign*}
\small
p^{\rm na}(y|x,S)=\frac{e^{\mathcal{D}(\mathcal{F}(x),r_y)}}{e^{\mathcal{D}(\mathcal{F}(x),r_y)}+\sum\limits_{k\in C_{t}\setminus \{y\}}e^{\mathcal{D}(\mathcal{F}(x),r_k)+m}}.
\end{flalign*}

The above naive additive margin loss assumes all classes should be equally far away from each other and thus add a fixed margin among all classes. In this way, this loss forces the embedding module $\mathcal{F}$ to extract more separable visual features for samples from different classes, which benefits the FSL. However, the fixed additive margin may lead to mistakes on test samples of similar classes, especially for the FSL where very limited number of labelled samples are provided in the novel classes.

\begin{figure*}[t]
\begin{center}
\includegraphics[width=0.83\textwidth]{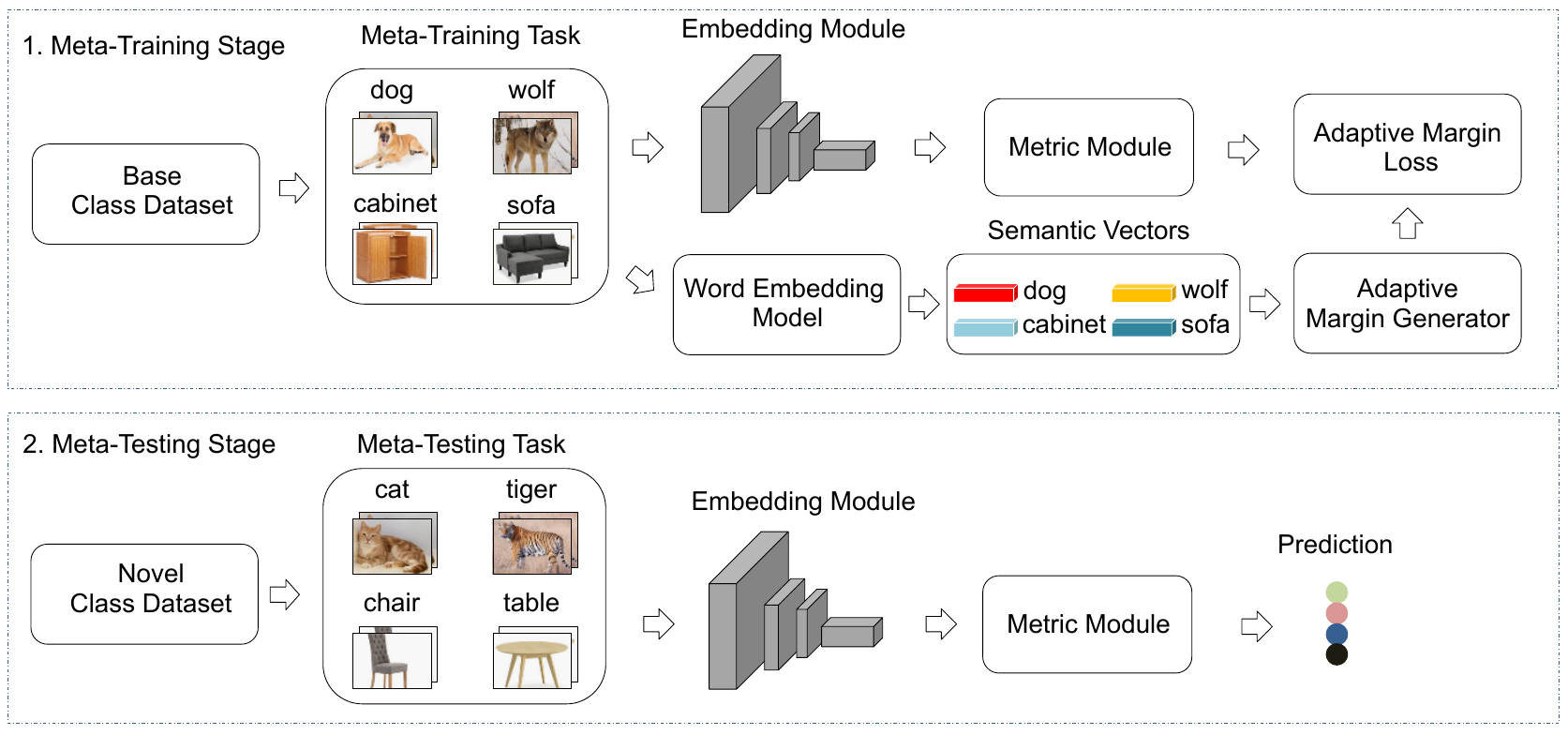}
\end{center} 
\caption{The overview of the proposed approach. Our approach consists of two stages: 1) In each episode of the meta-training stage,  we first sample a meta-training task from the base class dataset. Then, the names of classes in the meta-training task are fed into a word embedding model to extract semantic vectors for classes. After that, we propose an adaptive margin generator to produce margin penalty for each pair of classes (\eg, the class relevant margin generator proposed in Section \ref{sect:class_aware} or the task relevant margin generator proposed in Section \ref{sect:task_relevant}). Finally, we integrate the margin penalty into the classification loss and thus obtain an adaptive margin loss. A meta-learner consisting of an embedding module and a metric module is trained by minimizing the adaptive margin loss. 2) In the meta-testing stage, with the embedding module and metric module learned in the meta-training stage, we use a simple softmax (without any margin) to predict the labels of test novel samples. }
\label{overview}
\end{figure*}

\subsection{Class-Relevant Additive Margin Loss}
\label{sect:class_aware}

To better separate similar classes in the feature embedding space, the margin between two classes should be adaptive, \ie, the margin should be larger for similar classes than dissimilar classes. To achieve such adaptive margin in a principled manner, we design a class-relevant additive margin loss (CRAML), where semantic similarities between classes are introduced to adjust the margin. 

Before introducing the class-relevant additive margin loss, we first describe how to measure the semantic similarity between classes in a semantic space. Specifically, we represent each class name using a semantic vector extracted by
a word embedding model (\eg, Glove~\cite{glove}). As illustrated in Figure~\ref{overview}, we feed  a class name, such as wolf or dog, into the word embedding model, and it will embed the class name into the semantic space and return a semantic word vector. Then, we construct a class-relevant margin generator $\mathcal{M}$. For each pair of classes, class $i$ and class $j$, $\mathcal{M}$ uses their semantic word vectors $e_i$ and $e_j$ as inputs and generates their margin $m_{i,j}^{\rm cr}$ as follows:
\begin{equation}
\small
\begin{split}
m_{i,j}^{\rm cr} := \mathcal{M}(e_i,e_j)= \alpha \cdot \operatorname{sim}(e_i,e_j) +\beta,
\end{split}
\label{class_aware_margin_generator}
\end{equation}
where 
$\operatorname{sim}$ denotes a metric (\eg, cosine similarity) to measure the semantic similarity between classes.
We use $\alpha$ and $\beta$ to denote the scale and bias parameters for the class-relevant margin generator, respectively.

By introducing the class-relevant margin generator into the classification loss, we obtain a class-relevant additive margin loss as follows.
\begin{equation}
\small
\mathcal{L}^{\rm cr}\! =\!- \frac{1}{|Q|}\sum_{(x,y)\in Q}\log p^{\rm cr}(y|x,S),
\label{class_relevant_margin_loss}
\end{equation}
where 
\begin{flalign*}
\small
p^{\rm cr}(y|x,S)\!=\!\frac{e^{\mathcal{D}(\mathcal{F}(x),r_y)}}{e^{\mathcal{D}(\mathcal{F}(x),r_y)}+\sum\limits_{k\in C_{t}\setminus \{y\}}e^{\mathcal{D}(\mathcal{F}(x),r_k))+m_{y,k}^{\rm cr}}}.
\end{flalign*}

By exploiting the semantic similarity between classes properly, our class-relevant margin loss makes the samples from similar classes to be more separable in the embedding space. The more discriminative embedding space will help better recognize test novel class samples.

\begin{figure*}[t]
\begin{center}
\includegraphics[width=0.99\textwidth]{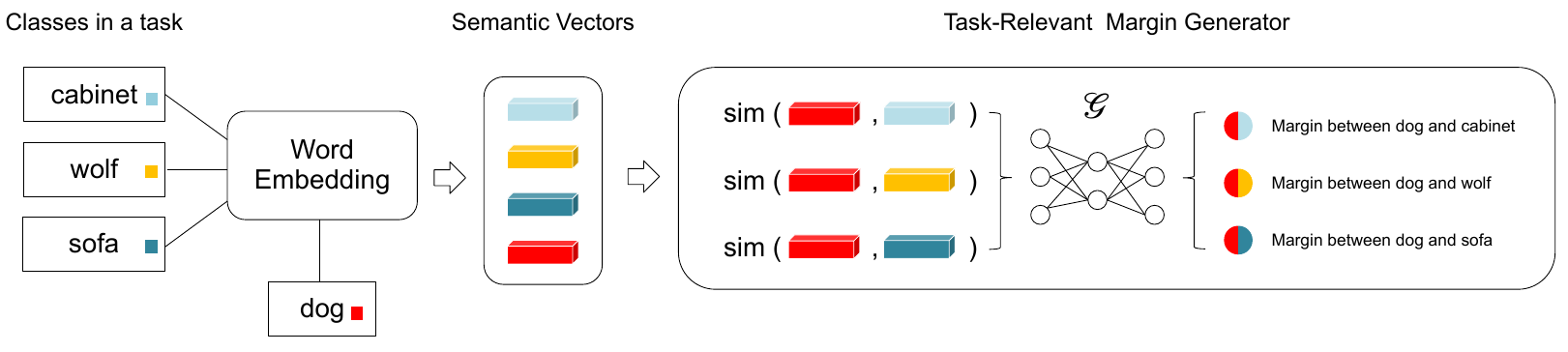}
\end{center}
\vspace{-0.1in}
\caption{The illustration of the architecture of our task-relevant margin generator.}
\label{arch_trm}
\end{figure*}

\subsection{Task-Relevant Additive Margin Loss}
\label{sect:task_relevant}

So far, we assume that the margin is task-irrelevant. A dynamic task-relevant margin generator, which considers the semantic context among all classes in a meta-training task, should generate more suitable margin between different classes. By comparing each class with other classes among a meta-training task, our task-relevant margin generator can measure the relatively semantic similarity between classes. Thus, the generator will add larger margin for relatively similar classes and smaller margin for relatively dissimilar classes. Therefore, we incorporate the generator into the classification loss and obtain the task-relevant additive margin loss (TRAML).

Specifically, given a class $y\in C_{t}$ in a meta-training task, the generator will produce the margins between class $y$ and the other classes $C_{t}\setminus \{y\}$ in the task according to their semantic similarities, namely,
\begin{equation}
\small
\{m_{y,k}^{\rm tr}\}_{k \in C_{t}\setminus \{y\}}= \mathcal{G}\left(\{\operatorname{sim}(e_{y},e_{k})\}_{k\in C_{t}\setminus \{y\}}\right),
\label{task_relevant_margin_generator}
\end{equation}
where $m_{y,k}^{\rm tr}$ denotes the task-relevant margin between class $y$ and class $k$, and $\mathcal{G}$ denotes the task-relevant margin generator, whose architecture is illustrated in Figure~\ref{arch_trm}. As shown in this figure, for a query sample (\eg, a dog image) with label $y \in C_{t}$, we first compute the similarities between its semantic vector $e_y$ and the semantic vectors of the other classes in the task (\eg, class wolf, sofa and cabinet), respectively. Then, these  semantic similarities\footnote{The order of input similarities has little impact on the performance.} are fed into the a fully-connected network to generate task-relevant margin for each class pair. By considering the context among all the classes in a meta-training task, our task-relevant margin generator can better measure the similarity among classes, thus generate more suitable margin for each class pair.

By integrating our task-relevant margin generator into the classification loss, we can obtain a task-relevant additive margin loss given in Equation~\ref{task_relevant_margin_loss} and the outline of computing task-relevant additive margin loss for a training episode is given in Algorithm~\ref{alg:traml}.
\begin{equation}
\small
\mathcal{L}^{\rm tr}\! =\!- \frac{1}{|Q|}\sum_{(x,y)\in Q}\log p^{\rm tr}(y|x,S),
\label{task_relevant_margin_loss}
\end{equation}
where 
\begin{flalign*}
\small
p^{\rm tr}(y|x,S)=\frac{e^{\mathcal{D}(\mathcal{F}\!(x),r_y)}}{e^{\mathcal{D}(\mathcal{F}(x),r_y)}+\sum\limits_{k\in C_{t}\setminus \{y\}} e^{{\mathcal{D}(\mathcal{F}(x),r_k)+m^{\rm tr}_{y,k}}}}.
\end{flalign*}

In a test episode, with the learned embedding module and metric module, we use the simple softmax function (without any margin) to predict the label of unlabeled data, \ie, we don't need to use semantic vectors of novel classes during the test stage, which makes our model flexible for any novel class.

\begin{algorithm}[t]
\caption{Task-relevant additive margin loss computation for a training episode in few-shot learning}
\label{alg:traml}
\textbf{Input:} Base class set $C_{base}$, task-relevant  generator $\mathcal{G}$.\\ 
\textbf{Output:} Task-relevant additive margin loss $\mathcal{L}^{\rm tr}$.

\begin{algorithmic}[1]
\STATE  Randomly sample $n_t$ classes from base class set $C_{base}$ to form an episodic training class set $C_{t}$;\\
\STATE  Randomly sample $n_s$ images per class in $C_{t}$ to form a support set $S$; \\
\STATE  Randomly sample $n_q$ images per class in $C_{t}$ to form a query set $Q$; \\
\STATE  Obtain the semantic vector for each class in $C_{t}$ by feeding its class name into a word embedding model; \\
\STATE  For each query sample, compute the task-relevant margins between its class $y$ and the classes in $C_{t}\!\setminus\!\{y\}$ by using task-relevant margin generator $\mathcal{G}$ according to Equation~\ref{task_relevant_margin_generator};\\
\STATE  Compute the task-relevant additive margin loss $\mathcal{L}^{\rm tr}$ according to Equation~\ref{task_relevant_margin_loss}.
\end{algorithmic}
\end{algorithm}

\subsection{Extension to Generalized Few-Shot Learning}

Although the proposed approach is originally designed for the standard FSL, it can be easily extended to the generalized FSL: simply including test data from both base and novel classes, and their labels are predicted from all classes in
both base and novel class set in the test stage. This setting is much more challenging
and realistic than the standard FSL, where test data are from only novel classes. Note that, our adaptive margin loss is flexible for the generalized FSL: the embedding module and the metric module trained by the adaptive loss with all training samples from base classes can be directly used for label inference of test samples from the disjoint space of both base and novel classes. Experimental results show that our method can improve the state-of-the-art alternative and create a new state-of-the-art for metric-based meta-learning approaches.

\begin{table*}[t]
\begin{center}
\tabcolsep0.3cm
\begin{small}
\begin{tabular}{lcccc}
\specialrule{0.05em}{0pt}{3pt}
\multirow{2}{*}{\bf Model}&\multirow{2}{*}{\bf Backbone}&\multirow{2}{*}{\bf Type}&\multicolumn{2}{c}{\bf Test Accuracy}\\
& &&5-way 1-shot & 5-way 5-shot \\\specialrule{0.05em}{2pt}{2pt}
Matching Networks \cite{vinyals2016bnips}& 4Conv&Metric& 43.56 $\pm$ 0.84 &55.31 $\pm$ 0.73\\
Prototypical Network \cite{Snell2017nips} & 4Conv&Metric&49.42 $\pm$ 0.78 &68.20 $\pm$ 0.66 \\
Relation Networks \cite{Sung2018cvpr} & 4Conv&Metric&50.44 $\pm$ 0.82 &65.32 $\pm$ 0.70 \\
GCR \cite{li2019few} & 4Conv &Metric&53.21 $\pm$ 0.40&72.34 $\pm$ 0.32\\
Memory Matching Network \cite{Cai2018cvpr}& 4Conv &Metric& 53.37 $\pm$ 0.48 & 66.97 $\pm$ 0.35 \\
Dynamic FSL \cite{Gidaris2018cvpr}& 4Conv&Metric& 56.20 $\pm$ 0.86&73.00 $\pm$ 0.64\\
Prototypical Network \cite{Snell2017nips}&ResNet12&Metric& 56.52 $\pm$ 0.45&74.28 $\pm$ 0.20\\
TADAM \cite{Oreshkin2018nips}& ResNet12&Metric& 58.50 $\pm$ 0.30&76.70 $\pm$ 0.38\\ 
DC \cite{Lifchitz2019cvpr} & ResNet12&Metric&62.53 $\pm$ 0.19&78.95 $\pm$ 0.13\\
TapNet \cite{Yoon2019icml} & ResNet12&Metric&61.65 $\pm$ 0.15&76.36 $\pm$ 0.10\\
ECMSFMT \cite{Ravichandran2019iccv} & ResNet12&Metric&59.00 &77.46 \\
AM3 (Prototypical Network) \cite{Chen2019NIPS} & ResNet12&Metric& 65.21$\pm$0.49&75.20 $\pm$ 0.36\\
\specialrule{0.05em}{2pt}{2pt}
MAML \cite{Finn2017icml}& 4Conv&Gradient& 48.70 $\pm$ 1.84 & 63.11 $\pm$ 0.92\\
MAML++ \cite{Antoniou2018iclr}& 4Conv&Gradient& 52.15 $\pm$ 0.26&68.32 $\pm$ 0.44\\
iMAML \cite{Rajeswaran2019nips}& 4Conv&Gradient& 49.30 $\pm$ 1.88& - \\
LCC \cite{Liu2019arxiv}& 4Conv&Gradient&54.6 $\pm$ 0.4&71.1 $\pm$ 0.4\\
CAML \cite{Jiang2019iclr}& ResNet12&Gradient&59.23 $\pm$ 0.99&72.35 $\pm$ 0.18\\
MTL \cite{Sun2019cvpr} & ResNet12&Gradient&61.20 $\pm$ 1.80&75.50 $\pm$ 0.80\\
MetaOptNet-SVM \cite{Lee2019cvpr} &ResNet12&Gradient&62.64 $\pm$ 0.61&78.63 $\pm$ 0.46\\
\specialrule{0.05em}{2pt}{2pt}
Prototypical Network + TRAML (OURS) & ResNet12&Metric& 60.31 $\pm$ 0.48 & 77.94 $\pm$ 0.57\\
AM3 (Prototypical Network) + TRAML (OURS) & ResNet12&Metric& \textbf{67.10} $\pm$ 0.52&\textbf{79.54} $\pm$ 0.60\\
\specialrule{0.05em}{2pt}{0pt}
\end{tabular}
\end{small}
\end{center}
\vspace{-0.0in}
\caption{Comparative results for FSL on the miniImageNet dataset. The averaged accuracy (\%) on 600 test episodes is given followed by the 95\% confidence intervals (\%). Notations: `4Conv' -- feature embedding module as in \cite{Snell2017nips}, \ie, four stacked convolutions layers of 64 filters; `ResNet12' -- the feature embedding module as in \cite{Oreshkin2018nips}, \ie, ResNet12 architecture containing four residual blocks of three stacked 3 $\times$ 3 convolutional layers; `Metric' -- metric-based meta-learning approaches for FSL; `Gradient' -- gradient-based meta-learning approaches for FSL.}
\label{fsl_miniImageNet}
\end{table*}

\section{Experiments and Discussions}

In this section, we evaluate our approach by conducting three groups of experiments: 1) standard FSL setting where the label space of test data is restricted to a few novel classes at each test episode, 2) generalized FSL setting where the label space of test data is extended to both base classes and novel classes, and 3) further evaluation including ablation study and comparison with other margin losses.

\subsection{Standard Few-Shot Learning}

\subsubsection{Datasets and Settings}

Under the standard FSL setting~\cite{Snell2017nips,vinyals2016bnips}, we evaluate our approach on the most popular benchmark, \ie,  miniImageNet. It contains 100 classes randomly selected from ImageNet \cite{Russakovsky2015ImageNet} and each class contains 600 images with resolution of 84 $\times$ 84. Following the widely used setting in prior works \cite{vinyals2016bnips,Snell2017nips}, we take 64 classes for training, 16 for validation and 20 for testing. In the training stage, the 64 training classes and 16 validation classes are respectively regarded as base classes and novel classes to decide the model hyperparameters. Following the standard setting adopted by most existing few-shot learning works \cite{vinyals2016bnips,Snell2017nips,Sung2018cvpr,Gidaris2018cvpr,Cai2018cvpr}, we conduct 5-way 1-shot/5-shot classification on the miniImageNet dataset. In 1-shot and 5-shot scenarios, each query set has 15 images per class, while each support set contains 1 and 5 image(s) per class, respectively. For a training episode, images in the support sets and query sets are randomly selected from the base class set. In a test episode, images in the support sets and the query sets are randomly selected from the novel class set. The evaluation metric for the miniImageNet dataset is defined as the top-1 classification accuracy on randomly selected 600 test episodes. We test our task-relevant additive margin loss with two backbone metric-based meta learning approaches:
Prototypical Networks \cite{Snell2017nips} and its most recent improvement AM3 (Prototypical Networks) \cite{Chen2019NIPS} which are the state-of-the-art  
metric-based meta learning methods for FSL. 

\begin{table*}[t]
\begin{center}
\tabcolsep0.3cm
\begin{small}
\begin{tabular}{l|ccccc|ccccc}
\specialrule{0.05em}{0pt}{3pt}
\multirow{2}{*}{\bf Model}&\multicolumn{5}{c|}{\bf Novel}&\multicolumn{5}{c}{\bf All}\\
& $n_s$=1&2&5&10&20&$n_s$=1&2&5&10&20 \\\specialrule{0.05em}{2pt}{2pt}
Logistic regression (from \cite{Wang2018cvprlsl}) &38.4&51.1&64.8&71.6&76.6&40.8&49.9&64.2&71.9&76.9\\
Logistic regression w/H (from \cite{Hariharan2017iccv}) &40.7&50.8&62.0&69.3&76.5&52.2&59.4&67.6&72.8&76.9\\
Prototypical Network \cite{Snell2017nips} (from \cite{Wang2018cvprlsl})& 39.3&54.4 &66.3&71.2&73.9&49.5&61.0&69.7&72.9&74.6 \\
Matching Networks \cite{vinyals2016bnips} (from \cite{Wang2018cvprlsl})& 43.6& 54.0&66.0&72.5&76.9&54.4&61.0&69.0&73.7&76.5\\
Squared Gradient Magnitude w/H \cite{Hariharan2017iccv}&-&-&-&-&-&54.3& 62.1&71.3&75.8&78.1\\
Batch Squared Gradient Magnitude \cite{Hariharan2017iccv}&-&-&-&-&-&49.3& 60.5&71.4&75.8&78.5\\
Prototype Matching Nets  \cite{Wang2018cvprlsl}&43.3&55.7&68.4&74.0&77.0&55.8&63.1&71.1&75.0&77.1\\
Prototype Matching Nets w/H \cite{Wang2018cvprlsl}&45.8&57.8&69.0&74.3&77.4&57.6&64.7&71.9&75.2&77.5\\
Dynamic FSL \cite{Gidaris2018cvpr}&46.0&57.5&69.2&74.8&78.1&58.2&65.2&72.2&76.5&78.7 \\
\specialrule{0.05em}{2pt}{2pt}
Dynamic FSL  + TRAML (OURS)&\textbf{48.1}&\textbf{59.2}&\textbf{70.3}&\textbf{76.4}&\textbf{79.4}&\textbf{59.2}&\textbf{66.2}&\textbf{73.6}&\textbf{77.3}&\textbf{80.2}
 \\
\specialrule{0.05em}{2pt}{0pt}
\end{tabular}
\end{small}
\end{center}
\vspace{-0.0in}
\caption{Comparative results for generalized FSL on the ImageNet2012 dataset. The top-5 accuracies (\%) on the novel classes and on all classes are used as the evaluation metrics for this dataset. Methods with “w/ H” use mechanisms that hallucinate extra training examples for the novel classes.}
\label{gfsl}
\end{table*} 

\subsubsection{Implementation Details}

 Our feature embedding module mirrors the ResNet12 architecture used by \cite{Oreshkin2018nips}, which consists of four residual blocks. Each block comprises three stacked 3 $\times$ 3 convolutional layers. Each block is followed by max pooling. We use the same feature extractor on images in both the support set and query set. The fully-connected network in the relation module consists of two fully-connected layers, each followed by a batch normalization layer and a ReLU non-linearity layer. The word embedding model we used in this paper is Glove~\cite{glove}.

\subsubsection{Experimental Results}
\label{sec: mini}
Table \ref{fsl_miniImageNet} provides comparative results for FSL on the miniImageNet dataset. We can observe that: 1) our approach significantly improve the performance of baseline models (\ie, Prototypical Network \cite{Snell2017nips} and AM3 (Prototypical Networks \cite{Chen2019NIPS}). This indicates that the proposed task-relevant additive margin loss can boost performance of metric-based meta-learning approaches very effectively. 2) Our approach clearly outperforms the state-of-the-art FSL model on both 5-way 1-shot and 5-way 5-shot settings, thanks to the discriminative feature embedding learned by the proposed task-relevant additive margin loss. 

\subsection{Generalized Few-Shot Learning}

\subsubsection{Dataset and Settings}

To further evaluate the effectiveness of our approach, we test our approach in a more challenging yet practical generalized FSL setting, where the label space of test data is extended to both base and novel classes. Following \cite{Hariharan2017iccv,Wang2018cvprlsl,Gidaris2018cvpr}, we conduct experiment on the large-scale ImageNet2012 dataset. This benchmark splits the 1000 ImageNet classes into 389 base classes and 611 novel classes; 193 of the base classes and 300 of the novel classes are used for cross validation and the remaining 196 base classes and 311 novel classes are used for the final evaluation (for more details we refer to \cite{Hariharan2017iccv,Wang2018cvprlsl}). 

As in \cite{Gidaris2018cvpr}, the embedding module we used is ResNet10 network that gets as input images of 224 $\times$ 224 resolution. We compare our model with several generalized FSL alternatives: Matching Networks \cite{vinyals2016bnips}, Prototypical Networks \cite{Snell2017nips}, Logistic Regression \cite{Wang2018cvprlsl}, Batch Squared Gradient Magnitude \cite{Hariharan2017iccv}, Squared Gradient Magnitude With Hallucination \cite{Hariharan2017iccv}, Prototype Matching Nets \cite{Wang2018cvprlsl}, and Dynamic FSL \cite{Gidaris2018cvpr}. 

We implement our task-relevant additive margin loss on the state-of-the-art model (\ie, Dynamic FSL~\cite{Gidaris2018cvpr}). Following \cite{Wang2018cvprlsl}, we first train the embedding module (\ie, ResNet10) by using our task-relevant additive margin loss with all base classes. Then we extract features for all training samples with the learned embedding module and save them to disk. The weight generator in Dynamic FSL \cite{Gidaris2018cvpr} will use these pre-computed features as inputs. Finally, we train the weight generator by replacing the original classification loss with our task-relevant additive margin loss. The evaluation metric is the top-5 accuracy on the novel classes and on all classes. We repeat the above experiment 5 times (sampling each time a different set of training images for the novel classes) and report the mean accuracy.

\subsubsection{Results}

Table \ref{gfsl} provides the comparative results of generalized FSL on the large-scale ImageNet2012 dataset. We can observe that: 1) our approach achieves the best results on all evaluation metrics. This indicates that, with the discriminative embedding space learned by our task-relevant additive margin loss, our approach has the strongest generalization ability under this more challenging setting. 2) Our approach yields consist performance improvement over the state-of-the-art generalized FSL model (\ie, Dynamic FSL \cite{Gidaris2018cvpr}) on the 1-shot, 2-shot, 5-shot, 10-shot, and 20-shot settings. This further validates the effectiveness of our approach. 

\subsection{Further Evaluation}

\subsubsection{Ablation Study on Key Components}
\label{ab}

 We compare our full model with a number of stripped down versions to evaluate the effectiveness of the key components of our approach. Specifically, three of such loss are compared, each of which uses the AM3 (Prototypical Networks) \cite{Chen2019NIPS} as the baseline model and differs only in which loss is used to train the model:  `Original Classification Loss' -- model training using the softmax loss provided in \cite{Chen2019NIPS}; `Naive Additive Margin Loss' -- model training by the loss proposed in Section~\ref{sect:simple_additive}; `Class-Relevant Additive Margin Loss' -- model training by the loss proposed in Section~\ref{sect:class_aware}. 
 
 Table \ref{ab_res} presents the comparative results of the above losses on the miniImageNet dataset under the standard FSL setting. It can be observed that: 1) Training metric-based meta-learning approaches with our adaptive margin loss leads to significant improvements (see Our Full Model vs. Original Classification Loss). This provides strong supports for our main contribution
on embedding learning for FSL. 2) The model trained by the proposed naive additive margin loss shows slight performance improvement over the model trained by original classification loss. This means that simply adding a fixed margin into the classification loss has limited effectiveness in FSL. 3) Thanks to the adaptive margin produced by the class-relevant margin generator, our class-relevant margin additive loss is shown to benefit the embedding learning for FSL (see Class-Relevant Additive Margin Loss vs. Naive Additive Margin Loss). 4) By considering the semantic context among classes in a meta-training task, our task-relevant additive margin loss yields better results than the class-relevant margin loss. 
Moreover, we observe that the learned coefficient $\alpha$ in Eq.~\eqref{class_aware_margin_generator} is positive, which verifies our intuition that the margin between similar classes should be larger than the one between dissimilar classes.

\begin{table}[t]
\begin{center}
\tabcolsep0.05cm
\begin{small}
\begin{tabular}{lcc}
\specialrule{0.05em}{0pt}{3pt}
\multirow{2}{*}{\bf Model}&\multicolumn{2}{c}{\bf Test Accuracy}\\
& 5-way 1-shot & 5-way 5-shot \\\specialrule{0.05em}{2pt}{2pt}
Original Classification Loss& 65.21 $\pm$ 0.49 & 75.20 $\pm$ 0.36\\
Naive Additive Margin Loss&  65.42 $\pm$ 0.25 & 75.48 $\pm$ 0.34\\
Class-Relevant Additive Margin Loss &  66.36 $\pm$ 0.57 & 77.21 $\pm$ 0.48\\
Our Full Model &\textbf{67.10} $\pm$ 0.52&\textbf{79.54} $\pm$ 0.60 \\
\specialrule{0.05em}{2pt}{0pt}
\end{tabular}
\end{small}
\end{center}
\vspace{-0.0in}
\caption{Ablation study for FSL on the miniImageNet dataset under the standard FSL setting. The evaluation metric is the same as in Table \ref{fsl_miniImageNet}.}
\label{ab_res}
\end{table}

\subsubsection{Comparison with Other Margin Losses}
\label{ab_2}

To validate the effectiveness of the proposed adaptive margin loss, we compare our
approach with two margin losses which are widely used in face recognition. Each of them uses the AM3 (Prototypical Networks) \cite{Chen2019NIPS} as the baseline model and differs only in which loss is used to train the model. The two margin losses are: 1) Additive angular margin loss \cite{Deng2019cvpr}, which add an additive angular margin to the angle between the weight vector and feature embeddings. 2) Additive cosine margin loss \cite{Wang2018cvprCosface}, which directly adds a cosine margin to the target logits. Note that, both of these two methods add margin penalty to the target logits computed by the dot product between feature embeddings and weight vectors. This is different from Prototypical Network and its variants, which use the opposite of the euclidean distances between class representations and feature embedding as the logits. For fair comparison, we replace the opposite of euclidean metric used in AM3 (Prototypical Network) \cite{Chen2019NIPS} with the cosine distance, and train the AM3 model with our task-relevant margin loss (the model is denoted by `Our Full Model (cosine)' in Table \ref{ab_res2}) . 

Table \ref{ab_res2} presents the comparative results of the two margin losses and our losses on the miniImageNet dataset under the standard FSL setting. We can observe that our method is shown to be more effective than the two competitors. It can be expected that, our method is designed for the FSL problem. That is, our method involves semantic similarity among classes in meta-training task to learn a more suitable margin penalty, compared with a fixed one generated by \cite{Deng2019cvpr,Wang2018cvprCosface}. The suitable margin of each pair of classes helps to learn more discriminative embedding space and thus better distinguish samples from different novel classes.

\begin{table}[t]
\begin{center}
\tabcolsep0.1cm
\begin{small}
\begin{tabular}{lcc}
\specialrule{0.05em}{0pt}{3pt}
\multirow{2}{*}{\bf Model}&\multicolumn{2}{c}{\bf Test Accuracy}\\
& 5-way 1-shot & 5-way 5-shot \\\specialrule{0.05em}{2pt}{2pt}
Additive angular margin loss \cite{Deng2019cvpr}&  66.21$\pm$0.46  & 77.30 $\pm$0.71 \\
Additive cosine margin loss \cite{Wang2018cvprCosface} & 65.96  $\pm$0.56  & 76.93 $\pm$0.49 \\
Our Full Model (cosine) &66.92 $\pm$ 0.43& 79.08 $\pm$ 0.52 \\
Our Full Model (euclidean) &\textbf{67.10} $\pm$ 0.52&\textbf{79.54} $\pm$ 0.60 \\
\specialrule{0.05em}{2pt}{0pt}
\end{tabular}
\end{small}
\end{center}
\caption{Comparative classification accuracies (\%) of two other margin losses on the miniImageNet dataset under the standard FSL setting. Notations: `Our Full Model (cosine)' -- implementing our task-relevant additive margin loss on AM3 (Prototypical Network) \cite{Chen2019NIPS} with cosine distance as metric in the embedding space; `Our Full Model (euclidean)' -- implementing our task-relevant additive margin loss on AM3 (Prototypical Network) \cite{Chen2019NIPS} with euclidean distance as metric in the embedding space. }
\label{ab_res2}
\end{table}

\section{Conclusion}

In this paper, we propose an adaptive margin principle, which can effectively enhance the discriminative power of embedding space for few-shot image recognition. We first develop a class-relevant additive margin loss which combines the standard classification loss with an adaptive margin generator based semantic similarity between classes. Then, by considering the semantic context among classes in a meta-training task, a task-relevant additive margin loss is further proposed to learn more discriminiative embbeding space for FSL. Furthermore, we also extend the proposed model to the more realistic generalized FSL setting. Experimental results demonstrate that our method is effective under both of the two FSL settings. 

\vspace{0.4cm}\noindent\textbf{Acknowledgment.} This work is supported by National Key R\&D Program of China (2018YFB1402600), BJNSF (L172037) and Beijing Acedemy of Artificial Intelligence.

\clearpage
{
\small
\bibliographystyle{ieee_fullname}
\bibliography{ama_bib}
}

\end{document}